\documentclass{article}

\PassOptionsToPackage{numbers, compress}{natbib}
\usepackage[preprint]{neurips_2025}

\usepackage[T1]{fontenc}    
\usepackage{hyperref}       
\usepackage{url}            
\usepackage{amsfonts}       
\usepackage{nicefrac}       
\usepackage{microtype}      
\usepackage{xcolor}         

\usepackage{booktabs} 

 \usepackage{amsmath}
 \usepackage{amssymb}
 \usepackage{mathtools}
 \usepackage{amsthm}
\usepackage{amsfonts}
\definecolor{darkgray1}{gray}{0.33}
\usepackage{colortbl}

\usepackage{xr}

\usepackage[capitalize,noabbrev]{cleveref}

\usepackage{bbm}
\newtheorem{theorem}{Theorem}[section]
\newtheorem{proposition}[theorem]{Proposition}

\newtheorem{definition}[theorem]{Definition}

\DeclareMathOperator{\E}{\mathbb{E}}

\DeclareMathOperator*{\argmax}{arg\,max}
\DeclareMathOperator*{\argmin}{arg\,min}
\DeclareMathOperator{\Conf}{Conf}

\theoremstyle{plain}
\theoremstyle{definition}

\theoremstyle{remark}

\usepackage[disable,textsize=tiny]{todonotes}

\title{Aligning Multiclass Neural Network Classifier Criterion with Task Performance Metrics}

\author{%
  Deyuan Li\footnotemark[2] \And Taesoo Daniel Lee\footnotemark[2] \And Marynel V\'{a}zquez\footnotemark[2] \And Nathan Tsoi\footnotemark[2] \\
  \AND \textnormal{\footnotemark[2]\,\,Yale University} \\
  \texttt{\{deyuan.li, taesoo.d.lee, marynel.vazquez, nathan.tsoi\}@yale.edu}
}

\begin{document}

\maketitle

\begin{abstract}
Multiclass neural network classifiers are typically trained using cross-entropy loss but evaluated using metrics derived from the confusion matrix, such as Accuracy, $F_\beta$-Score, and Matthews Correlation Coefficient. This mismatch between the training objective and evaluation metric can lead to suboptimal performance, particularly when the user's priorities differ from what cross-entropy implicitly optimizes. For example, in the presence of class imbalance, $F_1$-Score may be preferred over Accuracy. Similarly, given a preference towards precision, the $F_{\beta=0.25}$-Score will better reflect this preference than $F_1$-Score. However, standard cross-entropy loss does not accommodate such a preference. Building on prior work leveraging soft-set confusion matrices and a continuous piecewise-linear Heaviside approximation, we propose Evaluation Aligned Surrogate Training (EAST), a novel approach to train multiclass classifiers using close surrogates of confusion-matrix based metrics, thereby aligning a neural network classifier's predictions more closely to a target evaluation metric than typical cross-entropy loss. EAST introduces three key innovations: First, we propose a novel dynamic thresholding approach during training. Second, we propose using a multiclass soft-set confusion matrix. Third, we introduce an annealing process that gradually aligns the surrogate loss with the target evaluation metric. Our theoretical analysis shows that EAST results in consistent estimators of the target evaluation metric. Furthermore, we show that the learned network parameters converge asymptotically to values that optimize for the target evaluation metric. Extensive experiments validate the effectiveness of our approach, demonstrating improved alignment between training objectives and evaluation metrics, while outperforming existing methods across many datasets.
\end{abstract}

\section{Introduction}

Multiclass neural network classifiers are typically trained using cross-entropy loss but then evaluated using performance metrics based on the confusion matrix such as Accuracy, $F_1$-Score, or Matthews Correlation Coefficient (MCC).
This mismatch between the training objective and the evaluation metric can result in suboptimal performance, particularly when the evaluation metric reflects domain-specific priorities.
For example, when working with imbalanced data, Accuracy may be misleading, while metrics such as $F_1$-Score or MCC better capture the tradeoffs between precision and recall.
Likewise, when precision is more important than recall, the $F_{\beta=0.25}$-Score more accurately reflects performance than the standard $F_1$-Score. However, cross-entropy does not account for such nuances, leading to classifiers that are not optimized for the metrics that ultimately matter at evaluation.

A natural solution would be to use the final evaluation metric as the training criteria.
Indeed, prior work has explored using task-specific or metric-aligned surrogate losses for particular metrics in order to improve evaluation-time performance on these metrics~\cite{donti2017task, eban2017scalable, herschtal2004optimising, song2016training}.
However, it is generally infeasible to use a confusion-matrix based evaluation metric as the training objective when training via backpropagation, because they are computed on discrete prediction values and are therefore not continuous. Furthermore, because an $\argmax$ or Heaviside step function is used to assign predicted classes and covert network probability outputs into the confusion matrix sets, these functions have zero gradient almost everywhere and an undefined gradient at the decision threshold, making them unsuitable for gradient-based optimization methods.

To address this challenge, we propose Evaluation Aligned Surrogate Training (EAST), a novel training framework that enables neural networks to be optimized for a confusion-matrix based metric using a continuous and differentiable surrogate. EAST builds on prior work that introduced binary soft-set confusion matrices and a piecewise-linear approximation of the Heaviside function~\cite{tsoi2022bridging}.
EAST introduces three key innovations that enable optimizing multiclass neural network classifiers using close surrogates of confusion-matrix based evaluation metrics.
First, EAST introduces a dynamic thresholding mechanism that allows one to construct a continuous surrogate training loss whose gradient is suitable for backpropagation.
Second, we propose a multiclass soft-set confusion matrix, enabling EAST to support a broad range of metrics beyond binary classification.
Third, we propose the use of an annealing process that gradually sharpens surrogate losses towards the target evaluation metric, which slowly reduces any bias introduced when optimizing for the surrogate loss.

We provide a theoretical analysis showing that the EAST-based surrogates are consistent and asymptotically unbiased estimators of the given evaluation metric. We also prove that the network parameters converge to values that optimize for the intended metric. Empirical results across multiple datasets confirm that EAST achieves better alignment between training and evaluation objectives and in many cases outperforms baseline methods, including cross-entropy and recent surrogate approaches, on tasks where metrics like Macro $F_\beta$-Score, Accuracy, or MCC are indicative of network performance.

In summary, our three main contributions are (1) a new method for training multiclass neural networks using dynamic thresholding and annealing to optimize a continuous approximation of any confusion-matrix-based metric (Section~\ref{sec:method}), (2) a theoretical analysis demonstrating the consistency and convergence properties of EAST (Section~\ref{sec:theory}), and (3) extensive experiments showing EAST improves evaluation metric alignment and often outperforms prior methods across a range of settings (Section~\ref{sec:experiments}). 

\section{Related work}
\label{sec:related_work}

We are inspired by prior work on optimizing surrogate losses for binary classifiers~\cite{bao2020calibrated, koyejo2014consistent, tsoi2022bridging}, which laid the foundation for metric-aware learning. Additionally, annealing-based techniques have been explored to enable gradient-based learning in discrete settings, particularly through continuous relaxations of categorical distributions~\cite{jang2016categorical, maddison2016concrete}. Our work extends these ideas to the multiclass classification setting using neural networks, introducing a method for constructing training objectives that combine differentiable approximations with dynamic thresholding and annealing techniques.

Because backpropagation requires suitable gradients, a significant body of research has focused on constructing differentiable surrogates for non-differentiable evaluation metrics. These include $F_1$-Score~\cite{benedict2021sigmoidf1}, Dice~\cite{bertels2019optimizing, milletari2016v, sudre2017generalised}, Intersection over Union (IoU)~\cite{berman2018lovasz}, and Generalized IoU~\cite{rezatofighi2019generalized}. These surrogates enable optimization of task-specific objectives, especially in domains like medical imaging and object detection.
Our approach generalizes this line of work by enabling optimization for any confusion-matrix based metric, not just $F_\beta$-Score or Jaccard Index, through a novel combination of dynamic thresholding, a multiclass soft-set confusion matrix, and an annealing process~\cite{van1987simulated}. Our theoretical analysis in Section~\ref{sec:theory} shows that EAST-based surrogate losses are consistent and asymptotically unbiased estimators of a desired evaluation metric.

In settings with class imbalance, the choice of evaluation metric plays a critical role in evaluating classifier effectiveness~\cite{he2009learning, jeni2013facing, saito2015precision}. Metric-aware training has gained popularity for such scenarios, particularly through differentiable relaxations. For instance, \citet{milletari2016v} proposed soft Dice for binary segmentation tasks in medical imaging, and \citet{bertels2019optimizing} extended these ideas to Jaccard and $F_\beta$ in multi-label classification. EAST generalizes these techniques to a multiclass settings and enables optimizing a close surrogate of any confusion-matrix based metric.

While EAST is for neural networks optimized via backpropagation in a supervised regime, numerous alternative classification methods have been proposed, including Support Vector Machines (SVMs)~\cite{cortes1995support}, Random Forests (RFs)~\cite{ho1995random}, clustering-based techniques like $k$-means~\cite{hartigan1979algorithm}, adversarial formulations~\cite{fathony2020ap}, and linear classifiers in combination with Convex Calibrated Surrogates (CCS)~\cite{zhang2020convex}.
Although these approaches are well-established, they typically have less representational capacity than deep neural networks and often rely on two-stage pipelines involving separate training and post-hoc threshold selection~\cite{narasimhan2014statistical}.
In contrast, our approach integrates threshold selection directly into training by dynamically adapting the decision boundary throughout optimization. This eliminates the need for an additional tuning step and enables end-to-end optimization. By optimizing a differentiable surrogate of a target confusion-matrix-based evaluation metric, our method allows a neural network classifier to align its training objective directly with the metric used at evaluation time.

\section{Preliminaries}
\label{sec:preliminaries}
Multiclass classification involves determining the class membership of a given data sample among two or more possible classes, where each class is mutually exclusive.
Neural network-based classifiers are typically used to predict a probability distribution over the potential classes, where each data sample is ultimately assigned to one and only one class.
Formally, in a multiclass classification setting with $d$ classes, let such a neural network-based classifier have $d$ output nodes $\mathbf{z}=[z_1, ..., z_d]^\top$ where each $i$-th value in $\mathbf{z}$, for $1 \le i \le d$, corresponds to the likelihood that the input example belongs to the $i$-th class.
Typically, a softmax function is applied to $\mathbf{z}$ to obtain a probability vector, $\mathbf{p}$.
This vector represents the neural network's belief that a given output corresponds to the true label, where the $i$-th coordinate is $p_i = e^{z_i}/\sum_{j=1}^d e^{z_j}$. Let $\theta$ denote the values of the parameters of the neural network, and $f_\theta(x)$ denote the probability vector $\mathbf p$ that is outputted by the neural network on the input $x$ when using the parameters $\theta$.

Traditional methods train the neural network by minimizing the cross-entropy loss, constructed from the probabilities $p_i$, via backpropagation.
During evaluation, an input example is assigned the predicted class corresponding to $\hat y^H = \argmax_{1 \le i \le d} p_i$, where we use the notation $\hat y^H$ to distinguish between the soft-set predicted class vector, $\mathbf{\hat y}^{\mathcal{H}}$, that we introduce in Section \ref{sec:method}. 
The predicted class $\hat y^H$ is finally compared to the ground-truth label for the input to determine which possible confusion-matrix entry the prediction falls into, from which confusion-matrix based metrics can be computed.

In a multiclass setting with $d$ classes, a $d\times d$ multiclass confusion matrix $C \in \mathbb{R}_{\ge 0}^{d \times d}$ can be constructed. Rows correspond to the true classes while columns correspond to the predicted classes.
The entries of the multiclass confusion matrix consist of $\{c_{ij}\}_{1\le i,j \le d}$, where the $c_{ij}$ entry equals the number of total inputs with true class $i$ that are assigned a predicted label $j$ by the classifier. 

By treating a class in a one-versus-rest fashion, $2\times 2$ binary confusion matrices can be constructed from the multiclass confusion matrix. For instance, the entry in the binary confusion matrix for a given class $k$, where $k$ is the true class label, is:%
{\small
\begin{equation}
    \begin{split}
    |\mathit{TP}_k| = c_{kk}  \qquad |\mathit{FN}_k| = \sum_{i \neq k} c_{ki} \qquad \\
    |\mathit{FP}_k| = \sum_{i \neq k} c_{ik} \qquad |\mathit{TN}_k| = \sum_{i \neq k} \sum_{j \neq k} c_{ij}
    \label{eq:cardinalities}
    \end{split}
\end{equation}
}%
The entries of the class-specific confusion matrices are used to compute common classification metrics per class, from which a summary performance statistic can be derived.

For example, consider $F_\beta$-Score. Per-class results are typically combined by macro-averaging, where one first computes individual scores per class and then average the results. That is, for the $k$-th class, let $\text{Precision}_k = |TP_k|/(|TP_k| + |FP_k|)$ and $\text{Recall}_k = |TP_k|/(|TP_k| + |FN_k|)$. Then, $F_\beta$-Score is the weighted harmonic mean of precision and recall for that class, with $\text{$F_\beta$-Score}_k = (1+\beta^2)\frac{\text{Precision}_k\cdot \text{Recall}_k}{\beta^2\cdot\text{Precision}_k + \text{Recall}_k}$, and the macro-averaged score for all classes becomes:%
{\small
\begin{equation}
\text{Macro $F_\beta$-Score} = \frac{1}{d} \sum_{k=1}^d \text{$F_\beta$-Score}_k. 
\label{eq:macro-f-beta}
\end{equation}
}%
However, confusion-matrix based metrics such as Macro $F_\beta$-Score are not useful as a loss for training neural networks.
Recall that the network assigns an input example to the predicted class $\hat{y}^H = \argmax_{1 \le i \le d} p_i$, but this function is not continuous and has gradient $0$ everywhere it is defined.
Our proposed method, discussed in Section \ref{sec:method}, addresses this challenge by introducing a continuous approximation of the confusion matrix metric that can be optimized during training while maintaining theoretical guarantees.

\section{Method}
\label{sec:method}

Our approach enables training multiclass neural network classifiers to closely align with an evaluation metric by optimizing for a continuous and differentiable surrogate of the confusion-matrix based evaluation metric.
In order to accomplish this, our method incorporates three key components: dynamic thresholding, a view of the multiclass confusion matrix under soft-sets, and an annealing process.
The dynamic thresholding method and the multiclass soft-set confusion matrix are used to construct the surrogate loss at every epoch.
Each such training loss is parameterized by some temperature $T > 0$ that is slowly decreased following the annealing schedule.
We construct the method so that smaller temperature values correspond to closer approximations of the desired evaluation metric.
Therefore, as we progress through the annealing schedule, the training loss approaches the target evaluation metric, as shown in Section \ref{sec:theory}.

\subsection{Dynamic Thresholding}
\label{sec:piecewise-linear}
In order to compute a confusion-matrix based evaluation metric from the probability outputs ($\mathbf{p}$) of a neural network, one typically uses the $\argmax$ function to assign a predicted class.
Given the probability output $\mathbf p$, we propose a dynamic threshold $\tau_{\text{avg}}(\mathbf p)$ defined as the average of the two largest values of $\mathbf{p}$.
We chose this dynamic threshold so that the index of $\mathbf{p}$ whose probability value is greater than the dynamic threshold, is exactly equal to $\hat y^H = \argmax_i p_i$.
This is equivalent to applying the Heaviside step function $H$ at the threshold $\tau_{\text{avg}}(\mathbf p)$ coordinate-wise to each entry of $\mathbf p$: 
{\small
\begin{equation}
    \label{eq:one-hot-hard}
    \mathbf{\hat y}^H = H(\mathbf p, \tau_{\text{avg}}(\mathbf p)) = [H(p_1, \tau_{\text{avg}}(\mathbf p)), \ldots, H(p_d, \tau_{\text{avg}}(\mathbf p))]^\top.
\end{equation}
}%
We propose incorporating the dynamic threshold into a continuous approximation of $H$ whose gradients are useful for backpropagation. Inspired by \citet{tsoi2022bridging}, we propose a piecewise-linear approximation that incorporates the parameters necessary for both the dynamic thresholding as well as our annealing process. The Heaviside approximation we propose, $\mathcal{H}_T(p, \tau)$, is parameterized by some positive temperature $T \le 0.4$. In particular,
{\small
\begin{equation}
\label{eq:linear-param}
\mathcal{H}_T^l(p, \tau) =
  \begin{cases}
    p \cdot m_1 & \text{if $p < \tau - \frac{\tau_m}{2}$} \\
    p \cdot m_3 +(1-T-m_3(\tau+\frac{\tau_m}{2})) & \text{if $p > \tau + \frac{\tau_m}{2}$} \\
    p \cdot m_2 + (0.5 - m_2\tau) & \text{otherwise}
  \end{cases},
\end{equation}
}%
where $\tau_m = 5T\cdot \min\{\tau, 1-\tau\}$, $m_1 = T/(\tau - \frac{\tau_m}{2})$, $m_2 = (1-2T)/\tau_m$, and $m_3 = T/(1-\tau-\frac{\tau_m}{2})$. The construction of Equation \eqref{eq:linear-param} is described in Section \ref{sec:supp-prelim-approx}.

\subsection{Multiclass Soft-Set Confusion Matrix}
\label{sec:soft-set-confusion-matrix}
Our proposed approximation $\mathcal{H}_T^l$ and dynamic threshold $\tau_{\text{avg}}$ can then used to compute a continuous approximation of the predicted class.
Instead of associating a input $x$ with a singular predicted class, we relax this discreteness by assigning its membership to be a distribution over all classes via soft-sets~\cite{molodtsov1999soft}. The membership values should then sum to $1$ over all classes.
Therefore, we describe the degree to which an input example $x$ is assigned to each class via the function $g_T$. Given the probability distribution $\mathbf p = f_\theta(x)$, we let 
{\small
\begin{equation}
\label{eq:k_approx}
    g_T(\mathbf p) = \frac{\mathcal{H}_T^l(\mathbf{p}, \tau_{\text{avg}}(\mathbf{p}))}{\Vert\mathcal{H}_T^l(\mathbf{p}, \tau_{\text{avg}}(\mathbf{p}))\Vert_1} = \bigg[\underbrace{\frac{\mathcal{H}_T^l(p_1, \tau_{\text{avg}}(\mathbf{p}))}{\sum_{i=1}^d \mathcal{H}_T^l(p_i, \tau_{\text{avg}}(\mathbf{p}))}}_{(g_T(\mathbf p))_1}, \ldots , \underbrace{\frac{\mathcal{H}_T^l(p_d, \tau_{\text{avg}}(\mathbf{p}))}{\sum_{i=1}^d \mathcal{H}_T^l(p_i, \tau_{\text{avg}}(\mathbf{p}))}}_{(g_T(\mathbf p))_d}\bigg]^\top.
\end{equation}
}%
Then the $i$-th coordinate of $g_T(\mathbf p)$ equals the degree to which we assign the input to class $i$ via soft-sets.
Our method therefore approximates $\mathbf{\hat y}^H$ as $\mathbf{\hat y}_T^{\mathcal{H}} = g_T(\mathbf p)$.\footnote{The function $\mathcal{H}_T^l(\mathbf{p}, \tau_{\text{avg}}(\mathbf p))$ results in a continuous output, but there is no guarantee that $\sum_{i=1}^d \mathcal{H}_T^l(p_i, \tau_{\text{avg}}(\mathbf p)) = 1$.
Therefore, we additionally apply an $L_1$ normalization in Equation~\eqref{eq:k_approx}.}

The soft-set $d \times d$ confusion matrix values are computed by summing the class-wise probabilities into the appropriate entries. For an input with true label $i$, it contributes a total of $(\mathbf{\hat y}_T^{\mathcal{H}})_j$ into the $(i, j)$-th entry of the $d \times d$ confusion matrix for every $1 \le j \le d$. 

From these equations, we can construct soft-set versions of any confusion-matrix based evaluation metric.
Since each metric is some function $\mathcal{M}$ on the confusion matrix entries, the soft-set evaluation metric can be constructed by applying $\mathcal{M}$ to the continuous soft-set confusion matrix values.
This makes the soft-set confusion matrix and derived soft-set evaluation metrics continuous and provide a useful gradient for training multiclass neural network classifiers via backpropagation.

\subsection{Annealing Process}
Our method incorporates an annealing process that gradually aligns the surrogate loss with the target evaluation metric.
The annealing process we propose uses a geometric cooling schedule~\cite{kirkpatrick1983optimization} with temperatures $T_k = T_0 \cdot r^k$, where $0 < r < 1$ and $T_0$ is initialized to $0.2$.
The annealing process then proceeds as follows.
First, starting from $k = 0$, the soft-set metric is constructed from $g_{T_k}(\mathbf p)$ following the procedure in Section \ref{sec:soft-set-confusion-matrix}.
Second, the neural network is trained using the constructed soft-set metric via backpropagation at the current temperature $T_k$, and training continues until fast early stopping occurs~\cite{prechelt2002early}.
Then, we decrease the temperature from $T_k$ to $T_{k+1}$ and the early stopping counter is reset and the process repeats. Finally, the annealing process is stopped and training is complete when the validation loss stops decreasing across a number of temperature decreases.\footnote{Section \ref{sec:hyperparameters} details the selection of $T_0$ and the grid search we use to determine the other hyperparameters associated with the proposed annealing process.}

\section{Theoretical Grounding}
\label{sec:theory}
\subsection{Annealing Convergence Results}
\label{sec:conv-temp}
We provide a theoretical analysis of our approach to train neural networks for multiclass classification.
We show that the close surrogates of confusion-matrix based metrics computed via our method and used as a loss are continuous approximations that asymptotically converge to the true evaluation metric as $T \to 0$ along our cooling schedule.
\begin{definition}
\label{def: theta, g}%
Given a neural network with parameters $\theta$, let $f_\theta(x)$ denote the probability vector output of the neural network when given the input $x$. Also, for any probability vector $\mathbf p$, define 
\begin{equation*}
g(\mathbf p) = \mathrm{one\mbox{-}hot}(\argmax\nolimits_i p_i)
\end{equation*}
to be the one-hot-encoded vector representation of $\argmax_i p_i$.
\end{definition}
\vspace{-0.5em}
Note that under Definition \ref{def: theta, g}, $g(f_\theta(x))$ denotes the value $\mathbf{\hat y}^H$ from Equation \eqref{eq:one-hot-hard} assigned to input $x$ by the neural network. On the other hand, $g_T(f_\theta(x))$ represents the soft-set membership values $\mathbf{\hat y}_T^{\mathcal{H}}$. 

\begin{proposition} \label{prop: g-convergence}
For any $d$-dimensional probability distribution $\mathbf p$,
\begin{equation*}
\lim_{T \to 0} g_T(\mathbf p) = g(\mathbf p).
\end{equation*}
\end{proposition}
\begin{proof}
Recall from Equation~\eqref{eq:linear-param} that for $0 < T \le 0.4$,
{\small
\begin{equation*}
\label{eq:linear}
\mathcal{H}_T^l(p, \tau) =
  \begin{cases}
    p \cdot m_1 & \text{if $p < \tau - \frac{\tau_m}{2}$} \\
    p \cdot m_3 +(1-T-m_3(\tau+\frac{\tau_m}{2})) & \text{if $p > \tau + \frac{\tau_m}{2}$} \\
    p \cdot m_2 + (0.5 - m_2\tau) & \text{otherwise}
  \end{cases},
\end{equation*}
}%
where $\tau_m = 5T\cdot\min\{\tau, 1-\tau\}$ and $m_1 = T/(\tau - \frac{\tau_m}{2})$, $m_2 = (1-2T)/\tau_m$, $m_3 = T/(1-\tau-\frac{\tau_m}{2})$.
Let $\sigma \in S_d$ be the permutation from the $d$-dimensional symmetric group such that $p_{\sigma(1)} > p_{\sigma(2)} > \cdots > p_{\sigma(d)}$. Then for this input example, we dynamically threshold the Heaviside approximation $\mathcal{H}_T^l$ at $\tau_{\text{avg}}(\mathbf p) = \frac{p_{\sigma(1)} + p_{\sigma(2)}}{2}$. Thus, 
{\small
\begin{equation*}
p_{\sigma(1)} > \tau_{\text{avg}}(\mathbf p) > p_{\sigma(2)} > \cdots > p_{\sigma(d)}.
\end{equation*}
}%
When $T < \frac{2(p_{\sigma(1)}-p_{\sigma(2)})}{5}$ is sufficiently small, then
{\small
\begin{equation*}
\tau_m = 5T\cdot \min\{\tau_{\text{avg}}(\mathbf p), 1 - \tau_{\text{avg}}(\mathbf p)\} < p_{\sigma(1)}-p_{\sigma(2)},
\end{equation*}
}%
because $\min\{\tau_{\text{avg}}(\mathbf p), 1 - \tau_{\text{avg}}(\mathbf p)\} \le \frac{1}{2}$. In this case, $\frac{\tau_m}{2} < \frac{p_{\sigma(1)}-p_{\sigma(2)}}{2}$. It follows that $p_{\sigma(1)} > \tau_{\text{avg}}(\mathbf p) + \frac{\tau_m}{2}$, and $p_{\sigma(k)} < \tau_{\text{avg}}(\mathbf p) - \frac{\tau_m}{2}$ for all $k > 1$. Then for $k > 1$, by Equation~\eqref{eq:linear-param},
{\small 
\begin{equation*}
\lim_{T \to 0} \mathcal{H}_T^l(p_{\sigma(k)}, \tau_{\text{avg}}(\mathbf p)) = \lim_{T \to 0} p_{\sigma(k)}\cdot \frac{T}{\tau_{\text{avg}}(\mathbf p) - \frac{5T}{2}\cdot \min\{\tau_{\text{avg}}(\mathbf p), 1-\tau_{\text{avg}}(\mathbf p)\}} = 0.
\end{equation*}
}%
Also, 
{\small
\begin{equation*}
\begin{alignedat}{2}
    \lim_{T \to 0} \mathcal{H}_T^l(p_{\sigma(1)}, \tau_{\text{avg}}(\mathbf p)) &= \lim_{T \to 0} p_{\sigma(1)}m_3 + (1-T-m_3(\tau_{\text{avg}}(\mathbf p) + \tau_m/2)) \\
    &= 1,
\end{alignedat}
\end{equation*}
}%
because 
{\small
\begin{equation*}
\lim_{T \to 0} m_3 = \lim_{T \to 0} \frac{T}{1- \tau_{\text{avg}}(\mathbf p) - \frac{5T}{2}\cdot \min\{\tau_{\text{avg}}(\mathbf p), 1 - \tau_{\text{avg}}(\mathbf p) \}} = 0.
\end{equation*}
}%
Hence, 
{\small
\begin{equation*}
\lim_{T \to 0} \mathcal{H}_T^l(\mathbf p, \tau_{\text{avg}}(\mathbf p)) = H(\mathbf p, \tau_{\text{avg}}(\mathbf p)) = \mathrm{one\mbox{-}hot}(\sigma(1)) = g(\mathbf p).
\end{equation*}
}%
It also follows that 
{\small
\begin{equation*}
\lim_{T \to 0} \Vert \mathcal{H}_T^l(\mathbf p, \tau_{\text{avg}}(\mathbf p))\Vert_1 = \lim_{T \to 0} \sum_{i=1}^d \mathcal{H}_T^l(p_i, \tau_{\text{avg}}(\mathbf p)) = 1,
\end{equation*}
}%
and so 
{\small
\begin{equation*}
g_T(\mathbf p) = \frac{\mathcal{H}_T^l(\mathbf p, \tau_{\text{avg}}(\mathbf p))}{\Vert \mathcal{H}_T^l(\mathbf p, \tau_{\text{avg}}(\mathbf p))\Vert_1} \to g(\mathbf p)
\end{equation*}
}%
converges as $T \to 0$.
\end{proof}
\vspace{-0.5em}
Proposition \ref{prop: g-convergence} shows that for any input example with output probability distribution $\mathbf p$, the proposed method assigns soft-set membership values $g_T(\mathbf p)$ that converge to the true predicted class $g(\mathbf p)$ as $T \to 0$. Furthermore, for any set of examples $\{x_1, \ldots, x_n\}$, the $d \times d$ soft-set confusion matrix will consist of entries that converge to the true confusion matrix entries over the input set as shown in the following theorem.

\begin{theorem} \label{thm: metric-convergence}
Let $\mathcal{M} : \mathcal{C} \to \mathbb{R}$ be any evaluation metric that is continuous in the entries of the confusion matrix $C = (c_{ij})_{1 \le i, j \le 1} \in \mathcal{C}$, where $\mathcal{C} = \{C \in \mathbb{R}_{\ge 0}^{d \times d} : \Vert C \Vert > 0 \}$. 

Given a set of $n$ datapoints $(x_1, \ldots, x_n)$ with labels $(y_1, \ldots, y_n)$ and output probability distributions $(\mathbf p^{(1)}, \ldots, \mathbf p^{(n)}) = (f_\theta(x_1), \ldots, f_\theta(x_n))$, let $C$ be the confusion matrix constructed with $(g(\mathbf p^{(1)}), \ldots, g(\mathbf p^{(n)}))$ as the predicted classes, while let $C_T$ be the soft-set confusion matrix constructed with $(g_T(\mathbf p^{(1)}), \ldots, g_T(\mathbf p^{(n)}))$ as the corresponding predicted classes. Then as $T \to 0$,
$$C_T \to C \quad\text{and} \quad \mathcal{M}(C_T) \to \mathcal{M}(C).$$
\end{theorem}
\vspace{-0.5em}
\begin{proof}
From Proposition \ref{prop: g-convergence}, we know that for any $\mathbf p$, $g_T(\mathbf p) \to g(\mathbf p)$ as $T \to 0$. For the set of $n$ examples with known true classes of $(y_1, \ldots, y_n)$, let $\Conf : (\Delta^{d-1})^n \to \mathcal{C}$ denote the function that maps the predicted labels $\mathbf{\hat y}$ for the inputs to the corresponding confusion matrix. Here, 
{\small
\begin{equation*}
\Delta^{d-1} = \left \{(x^{(1)}, \ldots, x^{(d)}) \in \mathbb{R}_{\ge 0}^d: \sum_{i=1}^d x^{(i)} = 1\right \}
\end{equation*}
}%
is the $(d-1)$-dimensional probability simplex. Then $\Conf(g(\mathbf p^{(1)}), \ldots, g(\mathbf p^{(n)})) = C$ and $\Conf(g_T(\mathbf p^{(1)}), \ldots, g_T(\mathbf p^{(n)})) = C_T$. But since $\Conf$ is a continuous function, it follows from Proposition~\ref{prop: g-convergence} and the Continuous Mapping Theorem that $C_T \to C$ as $T \to 0$. Then since $\mathcal{M}$ is a continuous function over the entries of the confusion matrix,  by the Continuous Mapping Theorem, we also have that $\lim_{T \to 0} \mathcal{M}(C_T) = \mathcal{M}(C)$. 
\end{proof}
\vspace{-0.5em}
This implies that as $T \to 0$, the soft-set metric that we optimize over during training (computed from the soft-set confusion matrix entries) converges to the true evaluation metric. For bounded evaluation metrics (such as Macro $F_\beta$-Score, Accuracy, MCC, Precision, Recall), the Bounded Convergence Theorem guarantees that $\E[\mathcal{M}(C_T)] \to \mathcal{M}(C)$. 

Theorem \ref{thm: metric-convergence} implies that our proposed method optimizes over a surrogate loss that is an asymptotically consistent estimator of the true desired evaluation metric. This result suggests that our method of using soft-set confusion matrix entries to compute our evaluation metric $\mathcal{M}$ is a reasonable alternative to computing the true evaluation metric. 

\begin{theorem} \label{thm: weight-convergence}
Under mild assumptions, optimizing over the soft-set metrics via the proposed method yields network parameters that converge to the parameters that optimize the true evaluation metric.
\end{theorem}
\begin{proof}
Consider an annealing schedule $(T_k)_{k=0}^\infty$, where $T_0 = 0.2$ and the annealing schedule satisfies that $\lim_{k \to \infty} T_k = 0$ and $T_k > T_{k+1} > 0$ for all $k \ge 0$. Once again, for any $\mathbf p \in \Delta^{d-1}$, let $\sigma \in S_d$ be the permutation such that $p_{\sigma(1)} > \cdots > p_{\sigma(d)}$. If we further assume that $p_{\sigma(1)} - p_{\sigma(2)}$ is always bounded away from $0$, then the function $g_{T_k}(\mathbf p)$ converges uniformly to $g(\mathbf p)$ as $k \to \infty$. 

Let $\Theta$ be the set of all admissible trainable network parameters, which we consider to be a compact set. Then $\mathcal{M}(C_{T_k})$ and $\mathcal{M}(C)$ can be additionally viewed as functions over the trainable parameters $\theta$ of the neural network, and $\mathcal{M}(C_{T_k})$ converges uniformly to $\mathcal{M}(C)$ over $\Theta$. Suppose that $\theta^*$ corresponds to the set of parameters that uniquely minimizes our true desired evaluation metric $\mathcal{M}$. Then from Theorem \ref{thm: metric-convergence},
{\small
\begin{equation*}
\begin{alignedat}{2}
    \lim_{k \to \infty} \argmin_{\theta \in \Theta} \mathcal{M}(C_{T_k}) &= \argmin_{\theta \in \Theta} \lim_{k \to \infty} \mathcal{M}(C_{T_k}) \\
    &= \argmin_{\theta \in \Theta} \mathcal{M}(C) \\
    &= \theta^*,
\end{alignedat}
\end{equation*}
}%
which finishes the proof. 
\end{proof}
\vspace{-0.5em}
Theorem \ref{thm: weight-convergence} implies that minimizing the soft-set metrics $\mathcal{M}(C_T)$ via the proposed annealing schedule yields neural network parameters that converge to the parameters of the neural network that optimize the true evaluation metric, suggesting our method achieves the desired outcome.

\subsection{Guarantees on Dataset and Batch Size}
\label{sec:sub:dataset-size}
We present theoretical results for optimizing over a confusion-matrix based metric constructed from finitely many datapoints and compare how optimizing over a soft-set loss (e.g. soft-set Macro $F_\beta$-Score) constructed from a finite dataset differs from the population-level loss. We show that the empirical loss constructed from finitely many datapoints $n$ converges asymptotically to the population loss as $n \to \infty$. We also provide finite-sample guarantees and analyze this convergence rate. 
\begin{definition} \label{def: phi}
Let $\varphi : [d] \times \Delta^{d-1} \to [0,1]^{d \times d}$ be the function such that for any $1 \le i, j \le d$, the $(i, j)$-th entry of $\varphi(y, \mathbf{\hat y})$ is defined as
$$(\varphi(y, \mathbf{\hat y}))_{ij} = \begin{cases} \hat y_j & y = i \\ 0 & y \neq i \end{cases}.$$
\end{definition}

Definition \ref{def: phi} defines the function $\varphi$ which is used to assign an input $x$ with label $y$ and predicted class vector $\mathbf{\hat y}$ to the entries of the confusion matrix. Definition \ref{def: phi} is a generalization that holds for both soft-set and non-soft-set confusion matrices.
The confusion matrix constructed over a finite dataset is given by the sum of all the $\varphi(y, \mathbf{\hat y})$ values over each datapoint in the dataset.

\begin{definition} \label{def: confusion}
Suppose that $\mathcal{M} : \mathcal{C} \to \mathbb{R}$ is any continuous scale-invariant metric. Let $\theta$ be some set of parameters for the neural network and $g^* : \Delta^{d-1} \to \Delta^{d-1}$ be some function that maps probability output values to confusion-matrix membership vectors.

Suppose each datapoint $(X, Y)$ with input $X$ and label $Y$ is drawn from some overall population distribution $\mathcal{D}$. Then let
$$C_\theta = \E_{(X, Y) \sim \mathcal{D}}[\varphi(Y, g^*(f_\theta(X)))]$$
be the scaled population confusion matrix. For $n$ datapoints $(X_1, Y_1), \ldots, (X_n, Y_n)$ drawn i.i.d. from $\mathcal{D}$, let 
$$\hat C_{\theta, n} = \frac{1}{n} \sum_{i=1}^n \varphi(Y_i, g^*(f_\theta(X_i)))$$
be the empirical scaled confusion matrix over the dataset. 
\end{definition}

Note that in Definition \ref{def: confusion}, both $g_T$ and $g$ are examples of possible functions $g^*$. We then prove the following three theorems in Section \ref{sec:supp-theory}:

\begin{theorem} \label{thm: decomp-conv}
Let $\mathcal{M} : \mathcal{C} \to \mathbb{R}$ be any bounded, continuous evaluation metric that is additionally scale-invariant. Then for a finite set of input examples $(X_1, \ldots, X_n)$ with corresponding labels $(Y_1, \ldots, Y_n)$ randomly sampled from the overall population $\mathcal{D}$, 
$$\lim_{n \to \infty} \mathcal{M}(\hat C_{\theta, n}) = \mathcal{M}(C_\theta) \quad \text{and} \quad \lim_{n \to \infty} \E[\mathcal{M}(\hat C_{\theta, n})] = \mathcal{M}(C_\theta).$$
\end{theorem}

This shows that when the dataset is large enough, our proposed method enables optimization over an asymptotically unbiased and consistent estimator of the population metric.
Note that metrics such as Macro $F_\beta$-Score, Accuracy, MCC, Precision, and Recall all satisfy the conditions in Theorem~\ref{thm: decomp-conv}.
Even though these metrics are not decomposable, Theorem \ref{thm: decomp-conv} suggests that we can effectively train neural networks using our proposed surrogate losses because they are asymptotically unbiased estimators of the population metric as long as the batch size is sufficiently large.
\begin{theorem} \label{thm: concentration bound}
Consider a set of $n$ input examples $(X_1, \ldots, X_n)$ with corresponding labels $(Y_1, \ldots, Y_n)$ randomly sampled from the overall population $\mathcal{D}$, from which $\hat C_{\theta, n}$ is constructed via Definition \ref{def: confusion}. For any $0 < \delta < 1$, 
$$\Vert \hat C_{\theta, n} - C_\theta \Vert_{\max} \le \sqrt{\frac{\log(2d^2/\delta)}{2n}}$$
holds with probability at least $1-\delta$.
\end{theorem}
\begin{theorem} \label{thm: metric convergence rate}
Let $\mathcal{M} : \mathcal{C} \to \mathbb{R}$ be any continuous evaluation metric that is additionally scale-invariant. If $\mathcal{M}$ is also $L$-Lipschitz with respect to the entry-wise Chebyshev norm, then 
$$|\mathcal{M}(\hat C_{\theta, n}) - \mathcal{M}(C_\theta)| = O_p(L/\sqrt{n}).$$
\end{theorem}
\vspace{-0.5em}
Theorem \ref{thm: concentration bound} and Theorem \ref{thm: metric convergence rate} provide finite-sample guarantees on the deviations of the confusion matrix and the evaluation metrics when constructed from a finite dataset or batch size as opposed to the entire population $\mathcal{D}$. 

\section{Experiments}
\label{sec:experiments}
Our experiments show the effectiveness of EAST in optimizing three commonly used confusion-matrix based metrics: $F_1$-Score, Accuracy, and Matthews Correlation Coefficient (MCC) across six datasets including both binary and multiclass classification tasks. 
In Section \ref{sec:supp-experiments}, we further demonstrate EAST's flexibility by optimizing customized variants of these metrics, such as per-class $F_\beta$-Score where $\beta$ values encode preferences for precision or recall.
Full experimental details, including dataset descriptions, hyperparameter search ranges, fast early stopping criteria, and annealing process implementation are provided in Section \ref{sec:arch-train}.

\subsection{Results}
\label{sec:precision-recall-results}
\vspace{-0.1em}
As shown in Table \ref{tbl:results}, EAST consistently matches or outperforms the baselines on the target evaluation metric in nearly all settings.
For example when optimizing for $F_1$-Score (rows 1,6,11), EAST achieves higher $F_1$-Score than both cross-entropy and Dice loss. These results highlight EAST's ability to align training objectives with evaluation criteria more effectively than conventional approaches.

A notable exception occurs on the Kaggle binary classification dataset, composed of data with extreme class imbalance where only $0.17\%$ of examples are positive. In this case, cross-entropy slightly outperforms EAST when evaluated on $F_1$-Score. We attribute this to the difficulty of accurately estimating confusion-matrix based metrics over mini-batches of the data under such extreme imbalance. In practice, increasing batch size helps mitigate this issue (see Section~\ref{sec:hyperparameters} for details). However, increasing batch size introduces implicit regularization, which must be balanced with other hyperparameters. Although our hyperparameter search covered a reasonable range, further expanding the search space could improve performance in future work.

Our results also emphasize the importance of choosing an appropriate evaluation metric, particularly in imbalanced settings. For example, when predictions are imbalanced, MCC is known to produce wide fluctuations~\cite{grandini2020metrics}, making optimization challenging as we see with the Caltech256 dataset (row 3). Also, when optimizing for Accuracy on the Mammography and Kaggle datasets (row 12), EAST achieves strong Accuracy scores but performs poorly on $F_1$-Score. This is expected, as predicting only the dominant class in an imbalanced dataset can yield high Accuracy while ignoring the minority class entirely. In such cases, metrics like $F_1$-Score or MCC provide a more meaningful assessment of classifier performance~\cite{he2009learning}.

Additional experiments are included in Section \ref{sec:supp-experiments}, where we evaluate EAST on non-standard metrics that can be computed from the confusion-matrix. For example, $F_\beta$-Scores can be customized using a per-class $\beta$ value, which is useful in real-world cases where practitioners aim to prioritize precision or recall differently across classes. We compare EAST against a cross-entropy-trained classifier selected to match precision levels, demonstrating EAST’s ability to achieve the desired precision while still outperforming on the overall metric performance. We also include comparisons to additional baselines, such as random forest classifiers and linear classifiers trained with Convex Calibrated Surrogates (CCS) \cite{zhang2020convex}.

\begin{table*}[tb!p]
  \caption{
    Models trained on a range of multiclass (CIFAR-10, Caltech256) and binary (CocktailParty, Adult, Mammography, Kaggle) classification datasets and evaluated on $F_1$-Score, Accuracy (Acc), and Matthews Correlation Coefficient (MCC). Losses (rows) include Accuracy (\textit{Acc*}), $F_1$*, and MCC*, which use our proposed method, and baselines Dice~\cite{milletari2016v} and cross-entropy (\textit{CE}).
  }
  \centering
  \resizebox{\textwidth}{!}{
    \begin{tabular}{lrcccccc}
    &&\\
    & & \multicolumn{3}{c}{\textbf{CIFAR-10} ($\mu\pm\sigma$)} & \multicolumn{3}{c}{\textbf{Caltech256} ($\mu\pm\sigma$)} \\
    \cmidrule(lr){3-5} \cmidrule(lr){6-8}
    & \textit{Loss} & $F_1$ & Acc & MCC & $F_1$ & Acc & MCC \\
    \cmidrule(lr){2-2} \cmidrule(lr){3-3} \cmidrule(lr){4-4} \cmidrule(lr){5-5} \cmidrule(lr){6-6} \cmidrule(lr){7-7} \cmidrule(lr){8-8}
    (1) & $F_1$* &
    $0.765 \pm 0.01$ & $0.765 \pm 0.01$ & $0.738 \pm 0.01$ &
    $0.358 \pm 0.01$ & $0.367 \pm 0.01$ & $0.413 \pm 0.01$ \\
    (2) & \textit{Acc*} &
    $0.767 \pm 0.01$ & $0.768 \pm 0.01$ & $0.742 \pm 0.01$ &
    $0.299 \pm 0.01$ & $0.315 \pm 0.01$ & $0.370 \pm 0.01$ \\
    (3) & \text{MCC*} &
    $0.760 \pm 0.01$ & $0.761 \pm 0.01$ & $0.735 \pm 0.01$ &
    $0.024 \pm 0.01$ & $0.026 \pm 0.00$ & $0.214 \pm 0.01$ \\
    (4) & \textit{Dice} &
    $0.137 \pm 0.03$ & $0.154 \pm 0.03$ & $0.062 \pm 0.03$ &
    $0.001 \pm 0.00$ & $0.004 \pm 0.00$ & $0.003 \pm 0.01$ \\
    (5) & \textit{CE} &
    $0.755 \pm 0.01$ & $0.755 \pm 0.01$ & $0.728 \pm 0.01$ &
    $0.330 \pm 0.01$ & $0.333 \pm 0.01$ & $0.383 \pm 0.01$ \\
    &&\\
    & & \multicolumn{3}{c}{\textbf{CocktailParty} ($\mu\pm\sigma$)} & \multicolumn{3}{c}{\textbf{Adult} ($\mu\pm\sigma$)} \\
    \cmidrule(lr){3-5} \cmidrule(lr){6-8}
    & \textit{Loss} & $F_1$ & Acc & MCC & $F_1$ & Acc & MCC \\
    \cmidrule(lr){2-2} \cmidrule(lr){3-3} \cmidrule(lr){4-4} \cmidrule(lr){5-5} \cmidrule(lr){6-6} \cmidrule(lr){7-7} \cmidrule(lr){8-8}
    (6) & $F_1$* &
    $0.734 \pm 0.01$ & $0.844 \pm 0.01$ & $0.624 \pm 0.01$ &
    $0.364 \pm 0.10$ & $0.806 \pm 0.01$ & $0.369 \pm 0.05$ \\
    (7) & \textit{Acc*} &
    $0.724 \pm 0.01$ & $0.842 \pm 0.00$ & $0.614 \pm 0.01$ &
    $0.159 \pm 0.08$ & $0.786 \pm 0.01$ & $0.237 \pm 0.09$ \\
    (8) & \text{MCC*} &
    $0.724 \pm 0.01$ & $0.842 \pm 0.00$ & $0.616 \pm 0.01$ &
    $0.285 \pm 0.11$ & $0.797 \pm 0.01$ & $0.323 \pm 0.05$ \\
    (9) & \textit{Dice} &
    $0.299 \pm 0.15$ & $0.510 \pm 0.12$ & $-0.025 \pm 0.05$ &
    $0.246 \pm 0.14$ & $0.555 \pm 0.16$ & $0.023 \pm 0.09$ \\
    (10) & \textit{CE} &
    $0.730 \pm 0.01$ & $0.843 \pm 0.00$ & $0.626 \pm 0.01$ &
    $0.148 \pm 0.01$ & $0.782 \pm 0.00$ & $0.247 \pm 0.01$ \\
    &&\\
    & & \multicolumn{3}{c}{\textbf{Mammography} ($\mu\pm\sigma$)} & \multicolumn{3}{c}{\textbf{Kaggle} ($\mu\pm\sigma$)} \\
    \cmidrule(lr){3-5} \cmidrule(lr){6-8}
    & \textit{Loss} & $F_1$ & Acc & MCC & $F_1$ & Acc & MCC \\
    \cmidrule(lr){2-2} \cmidrule(lr){3-3} \cmidrule(lr){4-4} \cmidrule(lr){5-5} \cmidrule(lr){6-6} \cmidrule(lr){7-7} \cmidrule(lr){8-8}
    (11) & $F_1$* &
    $0.750 \pm 0.02$ & $0.989 \pm 0.00$ & $0.745 \pm 0.02$ &
    $0.824 \pm 0.01$ & $0.999 \pm 0.00$ & $0.824 \pm 0.01$ \\
    (12) & \textit{Acc*} &
    $0.748 \pm 0.02$ & $0.989 \pm 0.00$ & $0.745 \pm 0.02$ &
    $0.802 \pm 0.01$ & $0.999 \pm 0.00$ & $0.802 \pm 0.01$ \\
    (13) & \text{MCC*} &
    $0.787 \pm 0.03$ & $0.988 \pm 0.00$ & $0.784 \pm 0.03$ &
    $0.800 \pm 0.00$ & $0.999 \pm 0.00$ & $0.799 \pm 0.00$ \\
    (14) & \textit{Dice} &
    $0.077 \pm 0.08$ & $0.516 \pm 0.33$ & $0.063 \pm 0.12$ &
    $0.008 \pm 0.01$ & $0.479 \pm 0.30$ & $0.007 \pm 0.03$ \\
    (15) & \textit{CE} &
    $0.755 \pm 0.01$ & $0.990 \pm 0.00$ & $0.753 \pm 0.01$ &
    $0.855 \pm 0.01$ & $1.000 \pm 0.00$ & $0.856 \pm 0.01$ \\
    \end{tabular}
  }
  \label{tbl:results}
\end{table*}

\section{Limitations and Conclusion}
\label{sec:limitations-conclusion}
EAST offers a general and principled approach for aligning multiclass neural network classifiers with target evaluation metrics with broad applicability, as demonstrated on the datasets spanning domains including medical research, social signal processing, and image recognition.
However, our experiments were conducted on a limited number of datasets and our method applies only to metrics that are functions of the confusion matrix values.
Extending this work to include a wider range of task-specific metrics and additional real-world datasets is a promising direction for future work.
Given the potential for wide-ranging societal impact, we emphasize that practitioners should consider both the potential benefits and unintended side effects of improved classifier alignment in their specific application domain.

We have addressed the often overlooked misalignment between the training objective of a multiclass classification neural network and the confusion-matrix based metric used for evaluation.
Through the novel combination of dynamic thresholding during training, a multiclass soft-set confusion matrix, and an annealing process, EAST enables deep neural networks to be optimized for any confusion-matrix based metric, which should be chosen to align with application-specific performance goals.
Our theoretical analysis shows that EAST provides consistent estimators of a desired target evaluation metric and that, under mild assumptions, the network parameters converge asymptotically to optimal values under the surrogate objective.
Empirically, EAST outperforms standard cross-entropy loss across a range of binary and multiclass classification tasks.

\section{Acknowledgements}
\label{sec:ack}
This work was supported by the National Science Foundation (NSF), Grant No. (IIS-2143109). The findings and conclusions in this article are those of the authors and do not necessarily reflect the views of the NSF.

\bibliographystyle{plainnat}
\bibliography{references}

\newpage
\appendix
\section{Theoretical Grounding}
\label{sec:supp-theory}
This section provides the proofs for Theorem \ref{thm: decomp-conv}, Theorem \ref{thm: concentration bound}, and Theorem \ref{thm: metric convergence rate} mentioned in Section \ref{sec:sub:dataset-size} of the main paper.

\subsection{Guarantees on Dataset and Batch Size}
\setcounter{theorem}{6}
\renewcommand{\thetheorem}{5.\arabic{theorem}}
\label{sec:theory:dataset-size}

\begin{theorem}
Let $\mathcal{M} : \mathcal{C} \to \mathbb{R}$ be any bounded, continuous evaluation metric that is additionally scale-invariant. Then for a finite set of input examples $(X_1, \ldots, X_n)$ with corresponding labels $(Y_1, \ldots, Y_n)$ randomly sampled from the overall population $\mathcal{D}$, 
$$\lim_{n \to \infty} \mathcal{M}(\hat C_{\theta, n}) = \mathcal{M}(C_\theta) \quad \text{and} \quad \lim_{n \to \infty} \E[\mathcal{M}(\hat C_{\theta, n})] = \mathcal{M}(C_\theta).$$
\end{theorem}
\begin{proof}
Recall from Definition \ref{def: phi} that
$$C_\theta = \E_{(X, Y) \sim \mathcal{D}}[\varphi(Y, g^*(f_\theta(X)))]$$
and 
$$\hat C_{\theta, n} = \frac{1}{n} \sum_{i=1}^n \varphi(Y_i, g^*(f_\theta(X_i))).$$
But since the input examples $(X_k, Y_k) \sim \mathcal{D}$ are sampled i.i.d. from the population distribution $\mathcal{D}$, it follows from the Law of Large Numbers that
$$\hat C_{\theta, n} = \frac{1}{n} \sum_{i=1}^n \varphi(Y_i, g^*(f_\theta(X_i))) \to \E_{(X, Y) \sim \mathcal{D}}[\varphi(Y, g^*(f_\theta(X)))] = C_\theta$$
converges almost surely as $n \to \infty$. Then since $\mathcal{M}$ is a continuous function, it follows from the Continuous Mapping Theorem that
$$\lim_{n \to \infty} \mathcal{M}(\hat C_{\theta, n}) = \mathcal{M(C_\theta)}.$$
But $\mathcal{M}$ is also a bounded metric, so we therefore know from the Bounded Convergence Theorem that
$$\lim_{n \to \infty} \E[\mathcal{M}(\hat C_{\theta, n})] = \mathcal{M}(C_\theta).$$
\end{proof}

\begin{theorem}
Consider a set of $n$ input examples $(X_1, \ldots, X_n)$ with corresponding labels $(Y_1, \ldots, Y_n)$ randomly sampled from the overall population $\mathcal{D}$, from which $\hat C_{\theta, n}$ is constructed via Definition \ref{def: confusion}. For any $0 < \delta < 1$, 
$$\Vert \hat C_{\theta, n} - C_\theta \Vert_{\max} \le \sqrt{\frac{\log(2d^2/\delta)}{2n}}$$
holds with probability at least $1-\delta$.
\end{theorem}
\begin{proof}
For any $\delta \in (0, 1)$, consider the $(i, j)$-th entry of the confusion matrix. Let $\varphi_{ij}(y, g^*(f_\theta(x)))$ denote the $(i, j)$-th entry of $\varphi(y, g^*(f_\theta(x)))$. Then we know that for any $(x, y)$, 
$$0 \le \varphi_{ij}(y, g^*(f_\theta(x))) \le 1.$$ 
Thus, by Hoeffding's Inequality, for any $t > 0$,
$$\Pr\left(\left|\hat c_{ij,\theta, n} - c_{ij, \theta} \right| \ge t \right) \le 2 \exp(-2nt^2),$$
where $\hat c_{ij, \theta, n}$ and $c_{ij, \theta}$ denote the $(i,j)$-th entries of $\hat C_{\theta, n}$ and $C_\theta$, respectively. When $t = \sqrt{\frac{\log(2d^2/\delta)}{2n}}$, then 
$$\Pr\left(\left|\hat c_{ij,\theta, n} - c_{ij, \theta} \right| \ge t \right) \le 2 \exp(-2nt^2) = \frac{\delta}{d^2}.$$

Applying a union bound over all $d^2$ confusion matrix entries, it follows that
$$\Pr\left(\max_{i,j} \left|\hat c_{ij,\theta, n} - c_{ij, \theta} \right| \ge t \right) \le \delta,$$
and so 
$$\Vert \hat C_{\theta, n} - C_\theta \Vert_{\max} \le t = \sqrt{\frac{\log(2d^2/\delta)}{2n}}$$
holds with probability at least $1-\delta$.
\end{proof}

\begin{theorem}
Let $\mathcal{M} : \mathcal{C} \to \mathbb{R}$ be any continuous evaluation metric that is additionally scale-invariant. If $\mathcal{M}$ is also $L$-Lipschitz with respect to the entry-wise Chebyshev norm, then 
$$|\mathcal{M}(\hat C_{\theta, n}) - \mathcal{M}(C_\theta)| = O_p(L/\sqrt{n}).$$
\end{theorem}
\begin{proof}
Since $\mathcal{M}$ is $L$-Lipschitz with respect to the entry-wise Chebyshev norm, then
$$|\mathcal{M}(\hat C_{\theta, n}) - \mathcal{M}(C_\theta)| \le L \Vert \hat C_{\theta, n} - C_\theta\Vert_{\max}.$$
Using the result from Theorem \ref{thm: concentration bound}, it follows that for any $0 < \delta < 1$,
$$|\mathcal{M}(\hat C_{\theta, n}) - \mathcal{M}(C_\theta)| \le L\sqrt{\frac{\log(2d^2/\delta)}{2n}}$$
holds with probability at least $1 - \delta$. Thus, 
$$|\mathcal{M}(\hat C_{\theta, n}) - \mathcal{M}(C_\theta)| = O_p(L/\sqrt{n}),$$
since we consider the dimension $d$ to be fixed.
\end{proof}

\renewcommand{\thetheorem}{\thesection.\arabic{theorem}}
\section{Preliminaries}
\label{sec:supp-prelim}

\subsection{Linear Heaviside approximation}
\label{sec:supp-prelim-approx}

Recall the dynamic threshold that we propose and incorporate into a continuous approximation of $H$ whose gradients are useful for backpropagation.
Our piecewise-linear approximation of $H$, $\mathcal{H}_T^l(p, \tau)$,  incorporates the parameters necessary for both the dynamic thresholding as well as our annealing process.
This linear Heaviside approximation we propose is parameterized by some positive temperature $T \le 0.4$. From Equation \eqref{eq:linear-param} of the main paper, recall:

{\small
\begin{equation*}
\mathcal{H}_T^l(p, \tau) =
  \begin{cases}
    p \cdot m_1 & \text{if $p < \tau - \frac{\tau_m}{2}$} \\
    p \cdot m_3 +(1-T-m_3(\tau+\frac{\tau_m}{2})) & \text{if $p > \tau + \frac{\tau_m}{2}$} \\
    p \cdot m_2 + (0.5 - m_2\tau) & \text{otherwise}
  \end{cases}.
\end{equation*}
}%
As suggested by \citet{tsoi2022bridging}, the linear Heaviside approximation should meet the properties of a cumulative distribution function which ensures it is right-continuous, non-decreasing, with outputs in $[0,1]$ satisfying 
\begin{equation*}
\lim_{p \to 0} \mathcal{H}(p, \tau) = 0 \quad \text{and} \quad \lim_{p \to 1} \mathcal{H}(p, \tau) = 1
\end{equation*}
for all $\tau$. Then, we construct the five-point, linearly interpolated approximation $\mathcal{H}_T^l(p, \tau)$ similar to \citet{tsoi2022bridging}, which is defined over $[0,1]$ and is parameterized by the threshold $\tau$. However, uniquely in our work, we incorporate the temperature parameter $T$ by defining $\tau_m = 5T\cdot \min\{\tau, 1-\tau\}$.
Then, the three linear segments corresponding to slopes $m_1$, $m_2$, and $m_3$ that result in a function adhering to the desired properties are given by:

{\small
    \noindent\begin{minipage}{.33\linewidth}
    \centering
        $$
        m_1 = \frac{T}{\tau-\frac{\tau_m}{2}}
        $$
    \end{minipage}%
    \noindent\begin{minipage}{.33\linewidth}
    \centering
        $$
        m_2 = \frac{1-2T}{\tau_m}
        $$
    \end{minipage}%
    \noindent\begin{minipage}{.33\linewidth}
    \centering
        $$
        m_3 = \frac{T}{1-\tau-\frac{\tau_m}{2}}
        $$
    \end{minipage}
}%

Note that the constant $5$ in $\tau_m$ ensures that when $T = 0.2$, $\mathcal{H}_T^l(p, \tau)$ is equivalent to the $\mathcal{H}^l(p, \tau)$ function used in \citet{tsoi2022bridging}.

\section{Datasets}
\label{sec:data}

We performed experiments on multiclass classification datasets from various domains with different levels of class balance, as measured by Shannon's Equitability index~\cite{pielou1966measurement}.
We used two datasets in the domain of multiclass image classification. We also tested our methods against the four binary datasets proposed by \citet{tsoi2022bridging}.
For all datasets, we performed the minimal preprocessing steps of centering and scaling features to unit variance.
Each dataset was split into separate train (64\%), test (20\%), and validation (16\%) sets.

\subsection{Multiclass Datasets}
\label{sec:multiclass-datasets}

\textbf{CIFAR-10:}
The CIFAR-10 dataset\footnote{\url{https://www.cs.toronto.edu/ kriz/cifar.html}} consists of 60,000 images with even distribution amongst 10 classes (airplane, automobile, bird, cat, deer, dog, frog, horse, ship, and truck). Each image consists of three channels (RGB), with 32x32 pixels. Each class consists of 6,000 images. The images in CIFAR-10 are low-resolution, providing a suitable environment for testing an algorithm's ability to recognize patterns in small images. The Shannon's equitability index is 1, indicating a perfectly even distribution of images across the 10 classes. Note that a high Shannon equitability index indicates that a dataset has low class imbalance, and vice versa for low Shannon equitability index values. The \textit{Dog} class is at index 5 and the \textit{Frog} class is at index 6, which we use as examples in some of our experiments to evaluate how our method is capable of optimizing to balance between precision and recall.

\textbf{Caltech256:}
The Caltech256 dataset \cite{caltech256} comprises a collection of 30,607 images categorized into 256 object classes plus a ``clutter'' class, amounting to a total of 257 classes. Each class contains a minimum of 80 images, with the per-class total number of examples varying widely. The object categories range from animals and artifacts to human-made objects, offering a wide variety of image patterns for learning algorithms. The dataset, known for its high intra-class variability and degree of object localization within the images, often serves as a benchmark for object recognition tasks in computer vision. The class imbalance is moderate with a Shannon equitability index of 0.87. Note that the \textit{Dog} class is index 55 and the \textit{Frog} class is at index 79, which we use as examples in some of our experiments to evaluate how our method is capable of optimizing to balance between precision and recall.

\section{Binary Datasets}
\label{sec:binary-datasets}

We evaluated our proposed approach using the same four binary datasets proposed in prior work~\cite{tsoi2022bridging} to show the performance of our method on binary datasets with different levels of class imbalance.
The datasets are the CocktailParty dataset, which has a 30.29\% positive class balance~\cite{zen2010space}; the Adult dataset, which consists of salary data and has a 23.93\% positive class balance~\cite{Dua:2019}; the Mammography dataset, which consists of data on microcalcifications and has a 2.32\% positive class balance~\cite{woods1993comparative}; and the Kaggle Credit Card Fraud Detection dataset, which has a 0.17\% positive class balance~\cite{kaggle}.

\section{Neural Network Architecture and Training}
\label{sec:arch-train}

In order to fairly compare our proposed method of optimizing neural networks using a close surrogate of any confusion-matrix based evaluation metric with other baseline methods, we employed the same neural network architectures and training regimes across different datasets whenever possible.

\subsection{Protocol}
\label{sec:protocol}

We trained neural networks for each dataset using our proposed method to optimize for a close surrogate of common confusion-matrix based metrics including $F_1$-Score, Accuracy, Matthews Correlation Coefficient (MCC), and $F_\beta$-Score at different $\beta$ values that weight towards precision for a particular class or recall for a particular class.
We compared our method to baseline methods that utilize the same neural network architecture and training regime, but with other losses including Dice~\cite{milletari2016v} loss and typical cross-entropy loss. 
We also compared our method with non-neural network methods including Random Forests (RF)~\cite{ho1995random} and linear classifiers in combination with Convex Calibrated Surrogates (CCS)~\cite{zhang2020convex}.

Each dataset was segmented into training, validation, and testing splits.
Uniform network architectures and training protocols were applied wherever possible.
The AdamW optimizer~\cite{loshchilov2017decoupled} was used for training along with fast early stopping~\cite{prechelt2002early} up to 50 epochs without decreasing validation set loss, which we determined empirically.
Given the potential impact of hyperparameters on classifier performance, we performed a hyperparameter grid search for each combination of dataset and loss function.
We chose the hyperparameters that minimized validation-set loss and then performed 10 trials that varied the neural network random weight initialization. We then calculated the mean and standard deviation of the results across all trials. 

\subsection{Training Hardware and Software Versions}
\label{sec:training-hardware}

Training systems were equipped with a variety of NVIDIA general-purpose graphics processing units (GPGPUs) including Titan X, Titan V, RTX A4000, RTX 6000,  RTX 2080ti, and RTX 3090ti. Systems hosting these GPGPUs had between 32 and 256GB of system RAM and between 12 and 38 CPU cores. We used PyTorch with CUDA run inside a Docker container for consistency across training machines. 

\subsection{Architecture}

We employed two different model architectures which were chosen to align with the type of features in a particular dataset. One architecture was chosen for images datasets. Another architecture was utilized for tabular datasets. 

\subsubsection{Architecture for Image Data}

For image datasets, we used a convolutional neural network architecture that follows the design of Darknet-19, the backbone used in YoloV2~\cite{redmon2017yolo9000}. It is composed of $3 \times 3$ filtered and doubled channels after each pooling step. Filters of size $1 \times 1$ compress feature representations between $3 \times 3$ convolutions. Global average pooling is used for prediction.

\subsubsection{Architecture for Tabular Data}
\label{sec:tabular-arch}

For tabular datasets, we used a single fully-connected, feed-forward network consisting of four layers of 512 units, 256 units, 128 units, and $d$ units, where $d$ is the number of class labels in the dataset.
Rectified Linear Unit (ReLU) \cite{nair2010rectified} was applied at each intermediate layer, along with dropout \cite{hinton2012improving}.
We searched for the dropout hyperparameter value as described in Section \ref{sec:hyperparameters}.
The last layer of the neural network was linearly activated.

\subsection{Hyperparameters}
\label{sec:hyperparameters}

In order to fairly compare our proposed approach against the baselines, we searched for hyperparameter values using a limited grid search over batch size, learning rate, and regularization via dropout. Depending on the number of parameters in the network architecture and the features of the dataset, we searched over different ranges for batch size. For the Caltech256 dataset, we searched over batch size $\{32,64\}$. For the CIFAR-10 dataset, we searched over batch size $\{32, 64, 128, 256\}$.
For the binary datasets which are composed of tabular data, we searched over batch size $\{128, 256, 512, 1024, 2048\}$.
The one exception for the tabular dataset batch size search was the Kaggle dataset, which required a larger batch size for Accuracy and MCC loss due to the large class imbalance, as discussed in Section~\ref{sec:precision-recall-results} of the main paper, so we also additionally searched over $\{4069, 8192\}$.
Learning rate search was conducted over $\{0.01, 0.001, 0.0001\}$, and the dropout search range was $\{0.25, 0.5\}$.

We also searched for the annealing decay rate $r$ over the values $\{0.8, 0.9\}$. In our experiments, we initialize $T_0 = 0.2$ so that $\mathcal{H}_{T_0}^l$, described in Section~\ref{sec:supp-prelim-approx}, matches the approximation used by~\citet{tsoi2022bridging}.

\section{Experimental Results}
\label{sec:supp-experiments}

\subsection{Additional Image Dataset Baselines}
\label{sec:additional-baselines}

\begin{table}[tbp]
  \caption{
    Additional baseline results on the multiclass image datasets CIFAR-10 and Caltech256.
    We report the performance of neural networks trained using our method to optimize for $F_1$-Score ($F_1$*) and Accuracy (Acc*) loss. We compare these results to two baseline methods that do not use neural networks, a random forest (RF) and a linear classifier with Convex Calibrated Surrogates (CCS)~\cite{zhang2020convex}.
    Evaluation metrics are Macro $F_1$-Score and Accuracy. Each model was trained 10 times and we report the mean and standard deviation over these runs. For Caltech256 and CIFAR-10, we applied PCA before the RF and CCS methods to the image data, using only the requisite components to explain 80\% of the variance in the data. 
  }
  \centering
    \begin{tabular}{rcccc}
     & \multicolumn{2}{c}{\textbf{CIFAR-10} ($\mu\pm\sigma$)} & \multicolumn{2}{c}{\textbf{Caltech256} ($\mu\pm\sigma$)} \\
    \cmidrule(lr){2-3} \cmidrule(lr){4-5}
    \textit{Loss} &
      Macro $F_1$-Score & Accuracy &
      Macro $F_1$-Score & Accuracy \\
    \cmidrule(lr){1-1} \cmidrule(lr){2-2} \cmidrule(lr){3-3} \cmidrule(lr){4-4} \cmidrule(lr){5-5}
    \textit{$F_1$*} &
      $0.765 \pm 0.01$ & $0.765 \pm 0.01$ &
      $0.358 \pm 0.01$ & $0.367 \pm 0.01$ \\
    \textit{Acc*} &
      $0.767 \pm 0.01$ & $0.768 \pm 0.01$ &
      $0.299 \pm 0.01$ & $0.315 \pm 0.01$ \\
    \textit{RF} &
      $0.456 \pm 0.00$ & $0.460 \pm 0.00$ &
      $0.102 \pm 0.00$ & $0.180 \pm 0.00$ \\
    \textit{CCS} &
      $0.004 \pm 0.00$ & $0.111 \pm 0.00$ &
      $0.000 \pm 0.00$ & $0.005 \pm 0.00$ \\
    & \\
    \end{tabular}
  \label{tbl:additional-image}
\end{table}
\textbf{Experiment:} In addition to the experiments reported in the main paper, we compare the performance of our method to other baselines. In Table \ref{tbl:additional-image}, we compare the performance of models optimized for $F_1$-Score (\textit{$F_1$*}) and Accuracy (\textit{Acc*}) using our method against classical machine learning approaches that incorporate dimensionality reduction (PCA) with a random forest (RF)~\cite{ho1995random} or a linear classifier with Convex Calibrated Surrogates (CCS) as proposed by \citet{zhang2020convex}.

\textbf{Result:} On the multiclass image datasets, CIFAR-10 and Caltech256, the non-neural network approaches of Random Forest (RF) and Convex Calibrated Surrogates (CCS) underperform neural-network based methods, as reported in Table~\ref{tbl:additional-image}.
Neural networks optimized for $F_1$-Score and Accuracy outperform both the RF and CCS baselines, likely 
because neural networks excel at learning the high-dimensional representations typical of image datasets. Even though we made a substantial effort to make the RF and CCS baseline approaches perform better  by combining them with dimensionality reduction, they still fail to rival the performance of the neural network approaches.

\subsection{Additional Binary Dataset Baselines}
\label{sec:additional-binary-baselines}

\begin{table}[tbp]
  \caption{
    Additional baseline results on the binary datasets: CocktailParty, Adult, Mammography, and Kaggle. 
    We report the performance of neural networks trained using our method to optimize for $F_1$-Score ($F_1$*) and Accuracy (Acc*) loss. We compare these results to baseline methods that do not use neural networks, a random forest (RF)~\cite{ho1995random} and a linear classifier with Convex Calibrated Surrogates (CCS)~\cite{zhang2020convex}.
    Evaluation metrics are Macro $F_1$-Score and Accuracy. Each model was trained 10 times and we report the mean and standard deviation over these runs.
  }
  \centering
  \resizebox{\columnwidth}{!}{
    \begin{tabular}{lcccccccc}
    & \multicolumn{2}{c}{\textbf{CocktailParty} ($\mu\pm\sigma$)} 
    & \multicolumn{2}{c}{\textbf{Adult} ($\mu\pm\sigma$)} 
    & \multicolumn{2}{c}{\textbf{Mammography} ($\mu\pm\sigma$)} 
    & \multicolumn{2}{c}{\textbf{Kaggle} ($\mu\pm\sigma$)} \\
    \cmidrule(lr){2-3} \cmidrule(lr){4-5} \cmidrule(lr){6-7} \cmidrule(lr){8-9}
    \textit{Loss} 
    & $F_1$ & Acc 
    & $F_1$ & Acc 
    & $F_1$ & Acc 
    & $F_1$ & Acc \\
    \cmidrule(lr){1-1} \cmidrule(lr){2-2} \cmidrule(lr){3-3} \cmidrule(lr){4-4} \cmidrule(lr){5-5} \cmidrule(lr){6-6} \cmidrule(lr){7-7} \cmidrule(lr){8-8} \cmidrule(lr){9-9}\\
    \textit{$F_1$*} &
      $0.734 \pm 0.01$ & $0.844 \pm 0.01$ & 
      $0.364 \pm 0.10$ & $0.806 \pm 0.01$ & 
      $0.750 \pm 0.02$ & $0.989 \pm 0.00$ & 
      $0.824 \pm 0.01$ & $0.999 \pm 0.00$ \\
    \textit{Acc*} &
      $0.724 \pm 0.01$ & $0.842 \pm 0.00$ & 
      $0.159 \pm 0.08$ & $0.786 \pm 0.01$ & 
      $0.748 \pm 0.02$ & $0.989 \pm 0.00$ & 
      $0.802 \pm 0.01$ & $0.999 \pm 0.00$ \\
    \textit{RF} & 
      $0.794 \pm 0.01$ & $0.841 \pm 0.00$ & 
      $0.739 \pm 0.01$ & $0.836 \pm 0.00$ & 
      $0.854 \pm 0.01$ & $0.990 \pm 0.00$ & 
      $0.932 \pm 0.00$ & $0.999 \pm 0.00$ \\
    \textit{CCS} & 
      $0.738 \pm 0.00$ & $0.793 \pm 0.00$ & 
      $0.549 \pm 0.01$ & $0.553 \pm 0.01$ & 
      $0.799 \pm 0.00$ & $0.987 \pm 0.00$ & 
      $0.841 \pm 0.00$ & $0.999 \pm 0.00$ \\
    \end{tabular}
  }
  \label{tbl:additional-binary}
\end{table}

\textbf{Experiment:} In addition to the experiments on the binary datasets reported in the main paper, we provide additional baselines in Table \ref{tbl:additional-binary}. These experiments encompass two machine learning methods that do not utilize neural networks: a Random Forest (RF)~\cite{ho1995random} and linear classifier that incorporates Convex Calibrated Surrogates (CCS) proposed by \citet{zhang2020convex}.

\textbf{Result:} Non-neural network approaches perform well on the less complex binary datasets, but poorly on more complex, multiclass datasets as discussed in Section \ref{sec:additional-baselines}.
Notably the Random Forest (RF)~\cite{ho1995random} reliably outperforms all other methods on the tabular, binary datasets in Table~\ref{tbl:additional-binary}; however on the higher-dimensional multiclass image datasets in Table~\ref{tbl:additional-image}, performance is lacking.
Random Forests perform well on limited-complexity datasets due to their ensemble nature which can reduce overfitting even when data is scarce~\cite{breiman2001random}.
However, our method enables training multiclass neural network classifiers on more complex datasets using a close surrogate of the desired evaluation metric.
Moreover, these baselines do not enable optimization for any confusion-matrix based metric, as our method allows.

\subsection{Balancing Between Precision and Recall}
\label{sec:prec-recall-frog}
Suppose we are particularly concerned with a classifier performance on a specific class.
Our method is unique in that it makes it possible to train a multiclass neural network classifier using close surrogates of any confusion-matrix-based metric, including non-standard classification metrics.
For instance, if we prioritize precision or recall, our approach allows direct optimization of a surrogate to the  $F_\beta$-Score at any $\beta$ value. 

\textbf{Experiment}: In this additional experiment, we optimize for $F_\beta$-score. Given our method is capable of optimizing for a particular per-class preference towards precision or recall by adjusting the $\beta$ value per-class, we tested optimizing for either greater precision ($\beta=0.25$) or greater recall ($\beta=5$) for the \textit{Dog} class and then for the \textit{Frog} class. For all other classes, the per-class $\beta$ value was kept neutral ($\beta=1$), effectively making the per-class optimization objective $F_1$-Score for all classes except \textit{Dog} and \textit{Frog}. The $F$-Scores were computed for each class and all values were then macro-averaged. The \textit{Dog} and \textit{Frog} classes were chosen because they are both present in all of the multiclass datasets.

\textbf{Result:} Using our method, the neural network learns to output labels corresponding to increased precision or recall as directed during training while maintaining an overall Macro $F_1$-Score. 
Our method still outperforms the baseline cross-entropy loss, as shown in Table \ref{tbl:dog-and-frog}.
Networks trained to prefer precision are shown on the $F_\beta^P*$ lines, where we used a value of $\beta=0.25$.
Similarly, networks trained to prefer recall are shown on the $F_\beta^R*$ lines, where $\beta=5$ was used.
These results generalize across the example \textit{Dog} and \textit{Frog} classes.
Moreover, we also found that tuning $\beta$ for one class has minimal impact on the precision-recall tradeoff for another class.
Finally, we found that our method outperforms CE when using surrogate $F_\beta$ as a loss where $\beta$ is chosen to match the precision of the CE baseline. These results are detailed in Section \ref{sec:precision-recall-interplay}.

\begin{table*}[tb!p]
  \caption{
    Models trained to trade-off between precision and recall for the \textit{Dog} and \textit{Frog} classes. The $F_\beta^P*$ (where $\beta=0.25$) training criterion prefers precision and the $F_\beta^R*$ (where $\beta=5$) training criterion prefers recall. Results are reported for the Precision and Recall metrics on the \textit{Dog} and \textit{Frog} classes. We also report the Macro $F_1$-Score, which is $F_1$-Score averaged over all classes. Each model was trained 10 times and we report the mean and standard deviation over these runs.
  }
  \centering
  \resizebox{\textwidth}{!}{
    \begin{tabular}{rcccccc}
     & \multicolumn{3}{c}{\textbf{CIFAR-10} ($\mu\pm\sigma$)} & \multicolumn{3}{c}{\textbf{Caltech256} ($\mu\pm\sigma$)} \\
    \cmidrule(lr){2-4} \cmidrule(lr){5-7}
    \textit{Loss} &
      Precision (\textit{Dog}) & Recall (\textit{Dog}) &
      Macro $F_1$-Score &
      Precision (\textit{Dog}) & Recall (\textit{Dog}) &
      Macro $F_1$-Score  \\
    \cmidrule(lr){1-1} \cmidrule(lr){2-2} \cmidrule(lr){3-3} \cmidrule(lr){4-4} \cmidrule(lr){5-5} \cmidrule(lr){6-6} \cmidrule(lr){7-7} 
    $F_\beta^P*$ & 
      $0.789 \pm 0.04$ & $0.533 \pm 0.04$ & $0.754 \pm 0.01$ &
      $0.040 \pm 0.06$ & $0.038 \pm 0.05$ & $0.329 \pm 0.01$ \\
    $F_\beta^R*$ &
      $0.518 \pm 0.05$ & $0.769 \pm 0.03$ & $0.761 \pm 0.01$ &
      $0.048 \pm 0.04$ & $0.177 \pm 0.10$ & $0.362 \pm 0.01$\\
    \textit{CE} & 
      $0.655 \pm 0.07$ & $0.684 \pm 0.06$ & $0.746 \pm 0.01$ &
      $0.059 \pm 0.10$ & $0.054 \pm 0.08$ & $0.326 \pm 0.01$ \\
    \end{tabular}
  }

  \resizebox{\textwidth}{!}{
    \begin{tabular}{rcccccc}
     & \multicolumn{3}{c}{\textbf{CIFAR-10} ($\mu\pm\sigma$)} & \multicolumn{3}{c}{\textbf{Caltech256} ($\mu\pm\sigma$)} \\
    \cmidrule(lr){2-4} \cmidrule(lr){5-7}
    \textit{Loss} & Precision (\textit{Frog}) & Recall (\textit{Frog}) & Macro $F_1$-Score & Precision (\textit{Frog}) & Recall (\textit{Frog}) & Macro $F_1$-Score \\
    \cmidrule(lr){1-1} \cmidrule(lr){2-2} \cmidrule(lr){3-3} \cmidrule(lr){4-4} \cmidrule(lr){5-5} \cmidrule(lr){6-6} \cmidrule(lr){7-7} 
    $F_\beta^P*$ & $0.892 \pm 0.02$ & $0.714 \pm 0.03$ & $0.766 \pm 0.00$ &
      $0.124 \pm 0.15$ & $0.120 \pm 0.09$ & $0.370 \pm 0.02$ \\
    $F_\beta^R*$ & $0.611\pm 0.04$ & $0.879 \pm 0.02$ & $0.761 \pm 0.00$ &
      $0.036 \pm 0.03$ & $0.120 \pm 0.09$ & $0.363 \pm 0.01$\\
    \textit{CE} & $0.787 \pm 0.05$ & $0.800 \pm 0.06$ & $0.746 \pm 0.01$ &
      $0.022 \pm 0.06$ & $0.020 \pm 0.04$ & $0.326 \pm 0.01$\\
    \end{tabular}
  }
  \label{tbl:dog-and-frog}
\end{table*}

\subsection{Impact of Tuning $\beta$ for One Class on Another Class}
\label{sec:precision-recall-interplay}

\textbf{Experiment:} This experiment studies how the precision-recall tradeoff specified when training for one class affects the precision and recall performance of another class. To study these effects, we trained models as described in Section \ref{sec:arch-train} to optimize for a preference towards precision or recall for the \textit{Dog} class.
The network trained to prefer precision is shown on the $F_\beta^P*$ line, where we used a value of $\beta=0.25$.
Similarly, the network trained to prefer recall is shown on the $F_\beta^R*$ line, where $\beta=5$.
Along with the precision and recall performance of the \textit{Dog} class, we examine how precision and recall for the \textit{Frog} class changes. The \textit{Frog} and the \textit{Dog} class were selected because they are both present in both the CIFAR-10 and Caltech256 datasets.

\textbf{Result:} As reported in Table \ref{tbl:train-dog-only}, our method is effective at balancing towards the preferred precision or recall (\textit{Dog} class) while leaving the other class (\textit{Frog} class) precision vs. recall performance within a similar range.

\begin{table*}[tb!p]
  \caption{
    Models trained with a preference towards precision ($F^P_{\beta}*$ where $\beta=0.25$) or recall ($F^R_{\beta}*$ where $\beta=5.0$) for the \textit{Dog} class and evaluated on the precision and recall metrics for both the Dog class and the Frog class. Each model was trained 10 times and we report the mean and standard deviation over these runs.
  }
  \centering
  \resizebox{\textwidth}{!}{
    \begin{tabular}{rcccccccc}
     & \multicolumn{4}{c}{\textbf{CIFAR-10} ($\mu\pm\sigma$)} & \multicolumn{4}{c}{\textbf{Caltech256} ($\mu\pm\sigma$)} \\
    \cmidrule(lr){2-5} \cmidrule(lr){6-9}
    \textit{Loss} & Precision (\textit{Dog}) & Recall (\textit{Dog}) & Precision (\textit{Frog}) & Recall (\textit{Frog}) & Precision (\textit{Dog}) & Recall (\textit{Dog}) & Precision (\textit{Frog}) & Recall (\textit{Frog}) \\
    \cmidrule(lr){1-1} \cmidrule(lr){2-2} \cmidrule(lr){3-3} \cmidrule(lr){4-4} \cmidrule(lr){5-5} \cmidrule(lr){6-6} \cmidrule(lr){7-7} \cmidrule(lr){8-8} \cmidrule(lr){9-9}
    $F_\beta^P*$ & $0.831 \pm 0.03$ & $0.503 \pm 0.03$ & $0.837 \pm 0.03$ & $0.801 \pm 0.03$ &
      $0.078 \pm 0.10$ & $0.046 \pm 0.06$ & $0.043 \pm 0.04$ & $0.080 \pm 0.06$ \\
    $F_\beta^R*$ & $0.469 \pm 0.03$ & $0.810 \pm 0.03$ & $0.849 \pm 0.03$ & $0.776 \pm 0.03$ &
      $0.049 \pm 0.03$ & $0.185 \pm 0.13$ & $0.046 \pm 0.04$ & $0.090 \pm 0.07$  \\
    \end{tabular}
  }
  \label{tbl:train-dog-only}
\end{table*}
\subsection{A Neutral Classifier}
\label{sec:a-neural-network-classifier}

\begin{table}[ht]
  \caption{
    Models trained on the CIFAR-10 dataset to search for a $\beta$ value resulting in similar Precision to the CE baseline. Each model was trained 10 times and we report the mean and standard deviation over these runs.
  }
  \centering
  \begin{tabular}{rcc}
  $\beta$ & Precision & Recall \\
  \midrule
  $0.1$  & $0.719 \pm 0.03$ & $0.652 \pm 0.03$ \\
  $0.25$ & $0.686 \pm 0.04$ & $0.687 \pm 0.03$ \\
  $0.5$  & $0.695 \pm 0.03$ & $0.674 \pm 0.04$ \\
  $1.0$  & $0.684 \pm 0.03$ & $0.684 \pm 0.04$ \\
  $2.5$  & $0.656 \pm 0.04$ & $0.705 \pm 0.03$ \\
  $5.0$  & $0.640 \pm 0.03$ & $0.717 \pm 0.03$ \\
  $10.0$ & $0.620 \pm 0.04$ & $0.712 \pm 0.04$ \\
  \end{tabular}
  \label{tbl:neutral-model-beta-search}
\end{table}

 \begin{table}[ht]
  \caption{
    Models trained on the CIFAR-10 dataset with a $\beta$ value that results in similar precision as the CE baseline. Each model was trained 10 times and we report the mean and standard deviation over these runs.
  }
  \centering
    \begin{tabular}{rccc}
    Loss & Precision (Dog) & Recall (Dog) & Macro $F_1$-Score \\
    \midrule
    CE & $0.655 \pm 0.07$ & $0.684 \pm 0.06$ & $0.746 \pm 0.01$ \\
    $F_{\beta=2.5}*$ & $0.656 \pm 0.04$ & $0.705 \pm 0.03$ & $0.768 \pm 0.01$ \\
    \end{tabular}
  \label{tbl:neutral-model}
\end{table}

\textbf{Experiment:}
In order to better understand the performance of our proposed method relative to the cross-entropy baseline, we trained a neutral version of the proposed classifier where the $\beta$ value for the \textit{Dog} class was chosen to match the precision of the CE baseline using the balanced CIFAR-10 dataset. We first searched for a beta value resulting in precision similar to the CE baseline. Second, we trained a model using this beta value and evaluated the precision, recall, and Macro $F_1$-Score compared to the CE baseline.

\textbf{Result:} The $\beta$ value search, reported in Table \ref{tbl:neutral-model-beta-search}, revealed that $\beta=2.5$ resulted in a similar precision value compared to the CE baseline. 

Having found that the value $\beta=2.5$ results in similar Precision as the CE baseline, we then trained a new model using the close surrogate of $F_{\beta=2.5}$-Score via our method. Results in Table \ref{tbl:neutral-model} show that even when optimizing using an $F_\beta$-Score surrogate tuned to a similar level of precision as the CE baseline, our method allows the network to perform similarly or better than the CE baseline in terms of Precision, Recall and Macro $F_1$-Score.

\end{document}